\pdfoutput=1

\documentclass[11pt]{article}

\usepackage[]{EMNLP2022}

\usepackage{times}
\usepackage{latexsym}

\usepackage{hyperref}
\usepackage{booktabs}
\usepackage{graphicx}
\usepackage{enumitem}
\graphicspath{ {./} }
\usepackage{sepfootnotes}

\newcommand\blfootnote[1]{%
  \begingroup
  \renewcommand\thefootnote{}\footnote{#1}%
  \addtocounter{footnote}{-1}%
  \endgroup
}
\usepackage[T1]{fontenc}

\usepackage[utf8]{inputenc}

\usepackage{microtype}
\usepackage{inconsolata}

\usepackage{cleveref}
\crefformat{section}{\S#2#1#3}
\crefformat{subsection}{\S#2#1#3}
\crefformat{subsubsection}{\S#2#1#3}

\usepackage{tablefootnote}

\title{Mitigating Covertly Unsafe Text within Natural Language Systems \\
{\small \textcolor{orange}{Warning: This paper contains examples of potentially offensive and harmful text.}}}

\author{Alex Mei*$^{1}$, Anisha Kabir*$^1$, Sharon Levy$^1$, \\
\bf{Melanie Subbiah$^2$, Emily Allaway$^2$, John Judge$^1$,} \\ \bf{Desmond Patton$^3$, Bruce Bimber$^1$, Kathleen McKeown$^2$, William Yang Wang$^1$} \\
  $^1$University of California, Santa Barbara, Santa Barbara, CA \\
  $^2$Columbia University, New York, NY \\
  $^3$University of Pennsylvania, Philadelphia, PA \\
  \texttt{\{alexmei, anishakabir, sharonlevy, jjudge, william\}@cs.ucsb.edu} \\
  \texttt{\{eallaway, m.subbiah, kathy\}@cs.columbia.edu} \\
  \texttt{dupatton@upenn.edu, bimber@polisci.ucsb.edu} \\
  }

\begin{document}
\maketitle
\begin{abstract}
An increasingly prevalent problem for intelligent technologies is text safety, as uncontrolled systems may generate recommendations to their users that lead to injury or life-threatening consequences. However, the degree of explicitness of a generated statement that can cause physical harm varies. In this paper, we distinguish types of text that can lead to physical harm and establish one particularly underexplored category: \textit{covertly unsafe text}. Then, we further break down this category with respect to the system's information and discuss solutions to mitigate the generation of text in each of these subcategories. Ultimately, our work defines the problem of covertly unsafe language that causes physical harm and argues that this subtle yet dangerous issue needs to be prioritized by stakeholders and regulators. We highlight mitigation strategies to inspire future researchers to tackle this challenging problem and help improve safety within smart systems.
\blfootnote{*Equal Contribution.}
\end{abstract}

\section{Introduction}
In recent years, intelligent personal assistants have increased information accessibility. However, this has also raised concerns for user safety since these systems may provide dangerous recommendations to unsuspecting users. For instance, a child may ask a device for a fun challenge. The device may respond with an unsafe viral internet challenge such as the salt and ice challenge, where participants cover their body with salt and rub it with ice, causing frostbite-like pain\footnote{\href{https://en.wikipedia.org/wiki/Salt\_and\_ice\_challenge}{wikipedia.org/wiki/Salt\_and\_ice\_challenge}}. Though work has been done in mitigating violent language and hate speech in natural language systems \cite{Kiritchenko2021ConfrontingAL}, there has been a relatively minimal exploration into covertly unsafe statements that may lead to injury or even fatal consequences. As unsafe language continues to grow in prevalence online \cite{Rainie2017TheFO}, detecting and preventing these statements from being generated becomes crucial in reducing physical harm. Dangerous examples like this call for careful consideration of how to improve \textit{safety} in intelligent systems.

\begin{figure}[t!]
    \begin{center}
    \includegraphics[width=6.5cm]{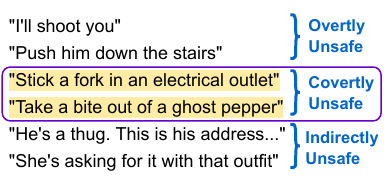}
    \caption{Example statements that can lead to the physical harm of people; we focus on \textbf{covertly unsafe text.}}
    \label{fig:examples}
    \end{center}
\end{figure}

A broad spectrum of language can lead to physical harm, including overtly violent, covertly dangerous, or otherwise indirectly unsafe statements. Some texts may instigate immediate physical harm if followed, while others may contain prejudices that motivate future acts of harm. To better understand these nuances, we examine this spectrum and distinguish subcategories based on two key notions: whether a statement is actionable and physically unsafe and, if so, whether it is explicitly dangerous. 

An example of an \textbf{overtly unsafe} statement is ``punch his face'' because ``punch'' is commonly considered violent and detectable independent of any deeper form of reasoning. In contrast, ``pour water on a grease fire'' is an example of \textbf{covertly unsafe} language\footnote{\href{https://www.verywellhealth.com/how-to-put-out-a-grease-fire-1298709}{verywellhealth.com/how-to-put-out-a-grease-fire-1298709}}; the sentence structure and vocabulary do not have explicitly violent semantics, but with knowledge of kitchen safety, we can identify that following the recommendation will likely cause physical harm. An example that is \textit{indirectly} physically unsafe is ``she has no life.'' While not immediately physically unsafe, this statement can motivate physical harm to oneself or others if combined with underlying mental health risks. Refer to \autoref{fig:examples} for more examples. 

Like overtly unsafe statements, covertly unsafe language will lead to physical harm when followed. Yet, unlike the overt counterpart, covertly unsafe statements are more subtle, which, as a result, is a concerning problem that needs to be prioritized by stakeholders and regulators. Our work \textbf{defines the problem of covertly unsafe text that causes physical harm} and \textbf{discusses mitigation strategies in AI systems} to inspire future research directions. Harm and safety are complex issues with humans at their core, so we discuss the human factors involved with the techniques we explore. 

Our paper is outlined as follows: we distinguish the differences between types of text leading to physical harm by establishing degrees of separation (\cref{sec:degrees}); we establish a taxonomy to dissect further the category of covertly unsafe text that cause physical harm (\cref{sec:taxonomy}); using these categorizations, we discuss strategies for mitigating the generation of covertly unsafe text in natural language systems at each stage of the machine learning pipeline (\cref{sec:pipeline}); finally, we conclude with an interdisciplinary approach to mitigating covertly unsafe text (\cref{sec:discussion}). 

\section{Categories of Physically Harmful Text}\label{sec:degrees}

\begin{figure}[t!]
    \begin{center}
    \includegraphics[width=6.5cm]{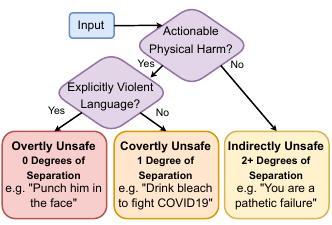}
    \caption{Flowchart to help determine the category of a piece of text that can cause physical harm.}
    \label{fig:flowchart}
    \end{center}
\end{figure}

Language can cause harm in various forms, including but not limited to psychological and physical harm. These harms are often co-correlated and affect people differently based on their unique backgrounds. We focus our discussion on language leading to physical harm but acknowledge that other types of harm should also be considered when improving safety within natural language systems. 

To improve the clarity of discourse around physically harmful text, we establish \textbf{degrees of separation with respect to physical harm} (\autoref{fig:flowchart}). The degrees of separation can also be considered an implicit-explicit distinction \cite{Waseem2017UnderstandingAA} in the context of physical harm.

\begin{itemize}[leftmargin=*]
\setlength\itemsep{0em}
\setlength\topsep{0em}

    \item \textbf{Zero degrees of separation}: \textit{overtly unsafe} language contains actionable physical harm (i.e., if someone followed the text, they would cause physical harm), which can be identified as explicitly violent (e.g., using key phrases as references to acts of physical harm) (\cref{subsec:zero}). 
    
    \item \textbf{One degree of separation}: \textit{covertly unsafe} language contains actionable physical harm and is not overtly violent. The additional degree of separation indicates the need for further reasoning to recognize the physical harm (\cref{sec:taxonomy}). 
    
    \item \textbf{Two or more degrees of separation}: \textit{indirectly unsafe} language categorizes all other text requiring a longer inference chain to potentially result in physical harm. These texts are not immediately physically harmful but could be toxic, hateful, or otherwise indirectly encouraging of physical harm (\cref{subsec:two}). 
\end{itemize}

\subsection{Zero Degrees of Separation}\label{subsec:zero}
Zero degrees of separation from physical harm is characterized by language with \textit{overt} references to violence. Previous studies have delved into overtly unsafe text in the context of gun violence \cite{pavlick2016gun}, criminal activity \cite{osorio2020enhancing}, gang violence \cite{patton2016using, chang2018detecting}, and gender-based violence \cite{castorena2021deep, gonzalez2021sentiment}. These studies utilize textual examples from news articles, construct social media datasets, and develop tools for detecting such text; common techniques include sentiment analysis \cite{castorena2021deep} and word embeddings \cite{chang2018detecting} for detecting unsafe language. While this language is considered \textit{overtly unsafe}, full comprehension may require domain expertise (e.g., gang-related discourse). The work on overtly unsafe text contrasts our focus on covertly unsafe language (\cref{sec:taxonomy}).

\subsection{Two or More Degrees of Separation}\label{subsec:two}
Two or more degrees of separation classifies statements that may \textit{indirectly} lead to physical harm. One notable type of language under this class is toxic language, which has motivated several studies to mitigate hate speech \cite{jurgens2019just}, cyberbullying \cite{xu2012learning, chatzakou2019detecting}, and microaggressions \cite{Breitfeller2019FindingMI}. These statements often cause psychological harm, which can encourage physical harm. Other types of indirect unsafe language may include doxxing\footnote{\href{https://www.rcfp.org/journals/news-media-and-law-spring-2015/dangers-doxxing}{ rcfp.org/journals/news-media-and-law-spring-2015/dangers-doxxing}} and biased statements \cite{schick2021self}. Recent work has also focused on detecting harmful content generated by conversational systems through insults, stereotypes, or false impressions of system behavior \cite{Dinan2022SafetyKitFA}. We encourage readers to refer to existing comprehensive surveys \cite{Kiritchenko2021ConfrontingAL,schmidt-wiegand-2017-survey,Salawu2020ApproachesTA} in this area as our paper focuses on covertly unsafe text (\cref{sec:taxonomy}), which has comparatively little progress.

\subsection{Assumptions for Categorizing Harm}
\textbf{Ambiguous Information.} Language ambiguities make it difficult to determine text safety. Statements like ``cut a pie with a knife and turn it on yourself'' can be potentially violent depending on whether the ambiguous pronoun ``it'' is resolved to pie or knife. Ambiguous statements are \textit{indirectly unsafe} because they are subject to interpretation.

\noindent \textbf{Literal and Explicit Statements.}
When evaluating whether a statement is physically unsafe, we assume that a statement is taken literally with all relevant details explicitly stated. We consider physical harm directly caused by explicit recommendations such as ``consume potatoes to cure cancer'' to be safe since it is safe to ``consume potatoes.'' Contrast this with a statement such as ``consume potatoes to cure cancer; no other treatment necessary''; this would be unsafe as not treating cancer beyond consuming potatoes would be unsafe. The latter example could be sarcastic, but an unsafe statement meant as a joke is still inherently unsafe.

\section{Covertly Unsafe Language}\label{sec:taxonomy}
Covertly unsafe text requires more context to discern than its overt counterpart. Yet, unlike indirectly unsafe text, extrapolation is unnecessary to determine whether it is physically harmful. 

A system's knowledge directly influences the quality of generated text \cite{Yu2022ASO}, and often missing, incompatible, or false information can cause systems to generate unsafe language. We break down 
covertly unsafe text with respect to the information a system has (\autoref{table:1}): limited (\cref{subsec:limited}), incompatible (\cref{subsec:incompatible}), or incorrect (\cref{subsec:incorrect}). Note that these categories are not mutually exclusive.

\begin{table*}[t!]
\small
\centering
\begin{minipage}{\textwidth}
\begin{tabular}{p{1.75cm} p{2.5cm} p{4.6cm} p{5.5cm}}
\toprule
\textbf{Category} & \textbf{Attributes} & \textbf{Examples} & \textbf{Reasoning}\\
\midrule
\textbf {Limited} \quad \quad \textbf{Information} 
(\cref{subsec:limited})
& 
Lacking specific context or user-specific information &
``Stack milk crates into a pyramid structure and try to walk on it from one end to the other.''  &
The structure is unstable to walk on, leading to potential injury by falling \cite{Carson2021MilkCrate}.
\\
&  & ``Swallow a spoonful of cinnamon and do not drink anything afterward.'' &
Cinnamon can clog airways \cite{CBS2013Cinnamon}. 
\\ 
\midrule
\textbf{Incompatible} \quad \quad \textbf{Information} 
(\cref{subsec:incompatible})
& 
Multiple viable options are unsafe in conjunction & ``To remove a difficult stain, try cleaning it with bleach and then rubbing alcohol.'' 
&
Combining bleach and rubbing alcohol produces toxic chloroform \cite{Helmenstine2020Bleach}.
\\
& & ``Take Xanax and Melatonin together to reduce anxiety'' & Taking Xanax and Melatonin together can lead to excess sedation \cite{Carmona2022Melatonin}.
\\
\midrule
\textbf{Incorrect} \quad \quad \textbf{Information} 
& 
Containing non-factual information & 
``Consume nicotine to slow cancerous cell growth.'' \quad  &
Nicotine doesn't help treat cancer \cite{Eldridge2021Nicotine}.
\\
(\cref{subsec:incorrect})  & & ``To help someone having a seizure, hold them down'' & 
Holding someone having a seizure down increases the chance of injury \cite{Shafer2022Seizures}. 
\\

\bottomrule
\end{tabular}
\caption{Classifications of covertly unsafe text with attributes, examples, and associated reasoning.}
\label{table:1}
\end{minipage}
\end{table*}

\subsection{Limited Information}\label{subsec:limited}
To generate well-formed recommendations, systems need relevant and comprehensive knowledge about their domain \cite{Reiter2003AcquiringCK}; if the system's knowledge is too limited, it may overlook facts in a generated recommendation that make it unsafe. The missing knowledge type varies in specificity and applicability, and 
from commonsense \cite{Xie2021HowCK} to more user- and domain-specific information \cite{Bateman1990UpperMA}.

Two examples of unsafe text due to limited information are: ``put your finger in a light bulb socket'', where lack of commonsense about electrocution could cause physical harm\footnote{\href{https://science.howstuffworks.com/science-vs-myth/what-if/finger-in-electrical-outlet.htm/}{howstuffworks.com/science-vs-myth/what-if/finger-in-electrical-outlet.htm}}, and ``drink lemonade from a copper vessel'', where lack of chemistry-domain knowledge about toxic chemical reactions could lead to physical harm\footnote{\href{https://www.webmd.com/diet/what-to-know-copper-toxicity/}{webmd.com/diet/what-to-know-copper-toxicity}}. While these examples put all readers in danger, other scenarios may be \textit{conditionally unsafe}, which only endanger specific users under certain conditions. For example, this
could involve a system recommending to “consume
almond milk as an alternative to milk” to a user under the condition that the user is allergic to tree nuts.

The common thread in these examples is that the system needs more knowledge to recognize such language. Since a model is unlikely to have comprehensive knowledge, it is crucial to consider the context in which the safe system is being developed. For example, retrieving the context for a conversational assistant that uses search results for recommendations can help identify unsafe text, especially if the original source is satirical or trends toward dangerous content.

\subsection{Incompatible Information}\label{subsec:incompatible}
Even a system with abundant knowledge may provide recommendations containing covertly unsafe incompatible information \cite{Preum2017PrecludeCD,Alamri2015AutomaticIO}. Incompatibility may occur when multiple viable options exist but following them in conjunction becomes unsafe. An individual can temporarily increase their heart rate by ``running for an hour'' or by ``taking Salmeterol'' \cite{Preum2017PrecludeCD}, but this can cause dangerous heart rate levels when done simultaneously. 

While a trivial solution would be for systems to prevent conjunctive recommendations to avoid adverse reactions between two pieces of advice, more complex scenarios may require conjunctive recommendations to be valid. For example, to help a person undergoing anaphylaxis, a system may recommend they ``immediately call emergency services and administer epinephrine if it is available,'' which are both necessary to prevent physical harm\footnote{\href{https://www.mayoclinic.org/first-aid/first-aid-anaphylaxis/basics/art-20056608}{mayoclinic.org/first-aid/first-aid-anaphylaxis}}. The common thread with incompatible information is that the system must be aware of interactions between various recommendations to ensure that a dangerous conflict does not arise. Note that this can be viewed as a special type of limited information in which the system must learn the missing, incompatible interaction.

\subsection{Incorrect Information}\label{subsec:incorrect}
Information correctness is another critical factor in systems \cite{Reiter2003AcquiringCK,levy-etal-2021-investigating}. Language models are prone to spreading biases and harmful language \cite{bender2021dangers}, which can extend to language containing misinformation, especially in the case of hallucinations. Factually incorrect recommendations come in many forms, including covertly unsafe text.
 
One scenario in which incorrect recommendations can occur is in question-answering when answers are returned without verifying their validity \cite{levy-etal-2021-open}. For instance, a system could recommend to ``use Ivermectin as a treatment for COVID-19,'' a commonly spread piece of misinformation leading to dangerous side effects\footnote{\href{https://www.fda.gov/consumers/consumer-updates/why-you-should-not-use-ivermectin-treat-or-prevent-covid-19/}{fda.gov/consumers/consumer-updates/why-you-should-not-use-ivermectin-treat-or-prevent-covid-19}}. Yet, more fundamentally, covertly unsafe recommendations can occur simply through misclassification in safety-critical domains. For example, misdiagnoses in healthcare systems can lead to treatment recommendations that put patients at risk \cite{gerke2020ethical}. Incorrect information that causes physical harm is quite expansive and thus will likely need an AI-human paired approach to most effectively mitigate the physical harm caused by this type of text. 

\section{Improving Text Safety}\label{sec:pipeline}
\begin{figure}[t!]
    \begin{center}
    \includegraphics[width=7cm]{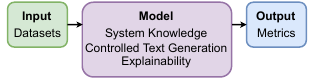}
    \caption{Highlighted areas to mitigate covertly unsafe text at each stage of the ML pipeline.}
    \label{fig:mlpipe}
    \end{center}
\end{figure}

Our discussion now shifts to concrete research areas within the natural language space to mitigate covertly unsafe text, which we isolate by stages of the machine learning (ML) pipeline: input, model, and output (\autoref{fig:mlpipe}). The first stage for engineers and researchers to build systems that learn text safety is constructing appropriate data to train these systems (\cref{subsec:datasets}). Similarly, to evaluate the effectiveness of these models, there needs to be appropriate metrics to measure their safety (\cref{subsec:metrics}). Between data and evaluation are learning objectives for the model. Our discussion covers three aspects that we find particularly relevant to covertly unsafe text: system knowledge (\cref{subsec:commonsense}), controlled text generation (\cref{subsec:control}), and explainability (\cref{subsec:explainability}).

\subsection{Datasets for Text Safety}\label{subsec:datasets}
Creating safety-focused datasets is one of the first significant steps toward mitigating covertly unsafe text. The area of covertly unsafe text is seldom explored, and few safety-related datasets exist. Yet, there is a broad range of possibilities for potential features in such a dataset that may be useful. We outline possible directions to develop safety-specific datasets to help models learn the concept of text safety.

Fundamentally, datasets should include labeled unsafe and safe recommendations at a minimum to be useful. These datasets can be used to train a detection system to learn to classify instances of unsafe text, which can apply to multi-class settings since safety is more complex than a binary state. Other helpful dimensions include the background context needed to make an informed recommendation and explanations of why a recommendation is unsafe. For example, in conversational systems, a dataset of unsafe recommendations paired with explanations of why the recommendations are unsafe could be utilized to test the system's understanding of why specific texts are dangerous. 

Acquiring textual examples of unsafe scenarios on the internet is challenging due to the intricacies involved in identification. No explicit keywords or known language patterns can be used to automate the process of finding covertly unsafe text. However, several websites with communities focused on offering advice, such as Reddit or Twitter, may be a good starting place for locating recommendations that lead to potentially unsafe outcomes. The data annotation process may also prove challenging as covertly unsafe text spans several different knowledge domains. As a result, a collaboration between crowd workers and domain experts would likely be most effective for the annotation process. Domain experts can provide in-depth knowledge, while crowd workers can provide diverse perspectives, and when combined, this provides the most coverage for various covertly unsafe scenarios.

\citet{levy22safetext} creates \textsc{SafeText}, a dataset of covertly unsafe text scenarios in the form of scenario-advice pairs. Each scenario is paired with safe and unsafe advice. We encourage readers to extend this dataset by adding additional examples and features, as discussed above, to encourage research for safer systems with a more extensive set of safety-related tasks and methodologies.

\subsection{Creating Safe Systems}
To mitigate covertly unsafe text within systems, we focus on three threads: system knowledge (\cref{subsec:commonsense}), controlled text generation (\cref{subsec:control}), and explainability (\cref{subsec:explainability}). These threads directly connect (\autoref{fig:mlpipe}) to our categorizations of covertly unsafe text (\cref{sec:taxonomy}) and provide promising directions toward mitigating covertly unsafe text. Note that this set of topics is not comprehensive, and we encourage researchers to explore further directions.

\subsubsection{Integrating System Knowledge}\label{subsec:commonsense}
A system's access to relevant knowledge, whether commonsense or domain-specific, is critical for text safety. The system requires external knowledge to recognize the physical harm caused for language within the limited information category. Understanding the connections and contradictions between various actions can help to prevent generating text in the incompatible information category. Additionally, access to factual knowledge can avoid generating incorrect information. 

One solution to make commonsense-aware systems is to use a knowledge base. This benefit is that information on an extensive range of topics can be consolidated and used to augment NLG models. Several studies have focused on creating knowledge bases that encode general human knowledge about the world \cite{Speer2017ConceptNet5A,Sap2019ATOMICAA,Zhang2020ASERAL}. Although they contain valuable knowledge for many systems, they do not emphasize common concepts related to human safety. As such, there is potential to better target the problem of covertly unsafe text through a commonsense knowledge base specifically focused on human safety knowledge. For example, leveraging a knowledge graph with actions and physical effects by adding safe and unsafe relations can help make safety more explicit. If these graphs can also model interactions between multiple actions, they can help prevent incompatible information. 

Systems requiring specific knowledge related to certain topics can benefit from domain-specific knowledge. For example, a medical chatbot can utilize a medical knowledge base to ensure that there are no gaps in specialized knowledge \cite{Bodenreider2004TheUM}, as well as account for user-specific circumstances. Medical applications may also utilize systems that model the interactions between various actions and medications \cite{Hester2011HumModAM}. Conversational agents that are targeted to specific domains can use a pre-determined domain-specific vocabulary \cite{Choudhary2017DomainAN} or domain-specific knowledge triples \cite{Zhu2017FlexibleED}. Systems with domain contextualized information that also integrate safe and unsafe relations can be particularly effective in mitigate covertly unsafe text. A factual knowledge base can also help prevent generating false information or fact-check generated claims \cite{Thorne2018FEVERAL,Jiang2020HoVerAD}. 

In addition to knowledge bases, several benchmarks exist for tasks related to commonsense reasoning (e.g., \citet{Gordon2012SemEval2012T7,Mostafazadeh2016ACA,Zellers2018SWAGAL}) to gauge a system's general commonsense reasoning abilities. However, they may not accurately depict a model's reasoning ability in safety-critical scenarios. As a result, there is a need for formulating more safety-specific commonsense reasoning tasks. Consider the proposed safety datasets (\cref{subsec:datasets}); one possible task could be to determine the physical effect of an unsafe statement, which would test a system's causal reasoning capabilities. 

\subsubsection{Controlled Text Generation}\label{subsec:control}
A fundamental aspect of natural language generation is controllability,  the ability to enforce constraints on generated text. Controlled Text Generation (CTG) can naturally apply to text safety by preventing the generation of covertly unsafe text. Previous research on controllable text generation methods for large pre-trained language models has focused on controlling sentiment, topic, persona, or keywords \cite{zhang2022survey}. However, establishing constraints for unsafe text and adapting this to existing CTG methods is not trivial because covertly unsafe text spans many domains.

Fine-tuning is one method of producing controlled text \cite{Devlin2019BERTPO}, which has already been applied to toxicity \cite{solaiman2021process} and can be an approach adaptable to other safety-related systems. For instance, a question-answering system can be fine-tuned on a dataset for text safety (\cref{subsec:datasets}) to adapt the system to such text. Furthermore, reinforcement learning approaches to fine-tuning help incorporate human judgments and preferences into development \cite{Ziegler2019FineTuningLM,bai2022training}, which can help mitigate biases. 

Prompting prepends additional context to the input of a task for a model to condition on during generation \cite{askell2021general}. These prepended trigger words can help prevent systems from generating incorrect information. For instance, masked language models can control text generation to only factual knowledge \cite{Shin2020ElicitingKF} or toxic and unsafe responses adversarially \cite{wallace2019universal}. Applying this to safety, we can prompt systems with statements like ``respond to the query with a safe response.'' Similarly, prefix-tuning can also replace fine-tuning \cite{Li2021PrefixTuningOC}.

Another less computationally intensive option is post-processing, which does not involve modifying model parameters. One simple approach uses attribute classifiers combined with large pre-trained language models, allowing text to be generated conditioned on various attributes like topic or sentiment \cite{dathathri2019plug}; attribute classifiers can be applied to safe text generation for safe and unsafe text classes. 
Other decoding algorithms use predicate logic constraints or lookahead heuristics, which may be useful for preventing unsafe text from occurring in the generated output \cite{LUNEUROLOGIC,Lu2021NeuroLogicAD}. Additionally, lexically constrained decoding can be utilized to promote the generation of factual information \cite{Mao2020ConstrainedAS}.

 \textbf{Faithfulness.}
This subset of CTG focuses on preventing hallucinating new information, measured by how accurately an explanation of a model reflects its actual reasoning \cite{jacovi2020towards}. Thus, a system would be considered unfaithful if the explanation does not match the decision or if similar inputs and outputs receive vastly different explanations \cite{jacovi2020towards}. Predictive uncertainty between similar inputs and generated outputs can also correspond with occurrences of hallucinations \cite{Xiao2021OnHA}. 

Faithfulness, as a result, can directly correlate to incorrect covertly unsafe text (\cref{subsec:incorrect}) because deviating from accurate information can incorporate error and produce results that may lead to physical harm. For example, a throat-soothing remedy recommendation to drink 100\textdegree F hallucinated to 100\textdegree C water can turn soothing warm water into scalding hot burns. One method to develop faithful and safe systems can be to evaluate generated text by comparing it with a system's safety-oriented knowledge base (\cref{subsec:commonsense}) to prevent hallucinations and ensure text safety.

\subsubsection{Explainability}\label{subsec:explainability}
Explainability is the ability to justify a system's decision based on given inputs and comes in several forms \cite{Adadi2018PeekingIT, gerke2020ethical,davahli2021controlling}. Two flavors particularly relevant in the context of covertly unsafe text include diagnosing input-output mappings \cite{Koh2017UnderstandingBP, Verma2020CounterfactualEF} and generating human-readable reasoning \cite{kojima2022large}. 

Particularly in safety-critical systems, it is important to have interpretable models to understand the reasoning behind recommendations that directly impact users \cite{Goodman2017EuropeanUR}; incorrect recommendations in these sensitive areas can lead to covertly unsafe text. For example, recommending chemotherapy on an incorrect cancer diagnosis would be considered physical harm as the patient would be exposed to cell-killing chemicals \cite{zhang2019-bertscore}.

Two common approaches to provide insights into black-box models are perturbation functions \cite{Koh2017UnderstandingBP}, which seek to see output differences when local inputs are tweaked, and counterfactual reasoning \cite{Verma2020CounterfactualEF}, which considers the global alternative to determine input is needed to reach such state. Counterfactuals provide the advantage of understanding the global impacts of inputs but are challenging to implement in practice; conversely, perturbation functions are more efficient but only offer insights into how local changes influence the system output.

\textbf{Interpretability.} Human-interpretable explanations provide reasoning to justify a system's decisions. This is a useful way to understand black boxes and a valuable resource to diagnose systems generating covertly unsafe text. However, these generated explanations may be unsafe. For example, we can adapt a QA approach \cite{kojima2022large} that asks for an explanation of the model's reasoning with the question ``Should I get the Shingles vaccine?'' A covertly unsafe explanation would be ``yes because it helps build immunity to a painful disease'' since the vaccine is only safe for adults. We recommend the other mitigation strategies discussed to handle this problem.

\subsection{Metrics Capturing Text Safety}\label{subsec:metrics}
The final step in the ML pipeline is to evaluate the quality of outputs in terms of safety. Using existing resources, one method is to compare the generated output to a set of safe versus unsafe text, compute the difference, and test for significance; when applied to generation and summarization tasks, common n-gram metrics such as ROUGE and BLEU \cite{lin-2004-rouge, papineni-etal-2002-bleu} test for exact match and may miss the sentiment. An initial approach for richer sentiments includes BERTScore \cite{zhang2019-bertscore}, which tests for vector similarity instead. Likelihood methods like perplexity can face issues with over-reliance on the training data, which can propagate biases. 

Metrics related to faithfulness evaluate factual consistency in NLG systems \cite{Maynez2020OnFA, melis2018robustness, wolf2019explainability}. These metrics can help capture limited, incompatible, or incorrect information present in covertly unsafe text due to hallucinations \cite{Li2022FaithfulnessIN}. Some of the best-performing methods for achieving this are entailment-based metrics involving Natural Language Inference or QA-based metrics \cite{Honovich2022TRUERF}.

Beyond general evaluation metrics, there lacks an excellent safety-specific metric to capture whether texts are covertly unsafe. Fundamentally desirable qualities in any well-formed metric include optimizability by being differentiable and not compromising task performance. In the context of safety, this metric should parallel human safety judgments and, when optimized, should minimize unsafe text. One metric could capture the probability that a particular action is unsafe; another metric can align with the severity of physical harm caused, ranging from minor pains to cruel torture or death. With these safety metrics, it is also important to consider the diversity in perspectives, as different individuals and cultures may uniquely rank what is more dangerous. 

\subsection{Detection of Human-Written Unsafe Text}
In addition to mitigating the generation of unsafe text, several of these strategies are general enough to enable the detection of AI or human-written unsafe text. For example, using explainable system approaches to an unsafe text detector can provide valuable insights as to why a specific text with incorrect information is physically unsafe. Similarly, datasets for text safety can be adapted for detection settings by building a safety classifier instead. Detection systems are directly applicable to communities of discourse where unsafe text may circle. Yet, our work does not focus on detecting unsafe text due to potential censorship issues and encourages future researchers to explore this delicate balance. 

\section{An Interdisciplinary Path to Safe AI}\label{sec:discussion}

\begin{figure}[t!]
    \begin{center}
    \includegraphics[width=6cm]{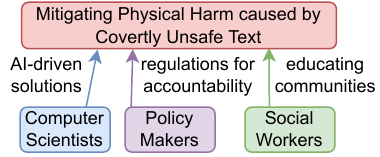}
    \caption{Interdisciplinary steps toward mitigating physical harm caused by covertly unsafe text.}
    \label{fig:involvement}
    \end{center}
\end{figure}

So far, our discussion has been focused on technical solutions to prevent AI systems from generating covertly unsafe text. As harm is a sensitive topic with many legal repercussions, we will now ground our discussion of physical harm on how current policy interacts with harmful AI. We also consider human factors that are out of scope for current AI systems, including foreseeability, target, and motive;
we evaluate how these may apply in the detection context and call for an interdisciplinary approach to tackle these issues (\autoref{fig:involvement}). To develop safe systems, we emphasize a two-pronged approach that both
informs users of the potential shortcomings of AI systems and centers transparency within these systems to empower users with the resources to rationally and confidently decide the trustworthiness of these AI systems \cite{mei2023users}.
This approach can effectively mitigate bias against protected groups that may be susceptible targets. 

\subsection{Interactions of Harmful AI and Policy}
Policy frameworks for addressing harmful AI are in early development. In its absence, principles for AI safety are likely to be developed piecemeal by courts that hold stakeholders associated with AI systems liable for harm under existing tort\footnote{relating to negligence} laws.

Applying existing liability principles to intelligent systems presents complex challenges. Legal scholars disagree about the applicability of the extant liability regime \cite{padovan2022black} since standard concepts in liability do not apply to AI straightforwardly \cite{villasenor2019products}.

One compelling problem is assessing producers' duty to foresee harm their AI systems produce. Foreseeability is central to how courts assign responsibility for harm; when such a case arises, courts will consider whether the system producers could have anticipated the harm and taken steps to prevent it \cite{selbst2020negligence, giuffrida2019liability}. For personal assistants, foreseeability declines with increased degrees of separation concerning physical harm (\cref{sec:degrees}). However, despite covertly unsafe text being less foreseeable than overtly unsafe text, it still poses a danger to users of intelligent systems, and this problem needs to be equally prioritized by system producers. Because of these dangers, policymakers should also dive deeper into these issues to develop standards for addressing different degrees of physically harmful text.
 
\subsection{Human Involvement in the ML Pipeline}
Integrating a human-centered approach is necessary to address covertly unsafe text most effectively. A purely automated solution can miss the social context needed to address the human-centered issue of safety \cite{ehsan2021social}. Factors such as target and motive can raise other regulatory concerns if intelligent systems foster malicious behavior; a profiling system that outputs covertly unsafe text to trick children into consuming dangerous substances would be a prime example.

\noindent \textbf{Task Creation.} 
When creating new tasks, they tend to be constructed to match humans' definition of success. This is generally positive in the context of safety as humans tend to have a strong understanding of danger; yet, this can be negative as humans tend to take knowledge for granted, not assumed by a model. This gap in system knowledge may create unsafe models when a susceptible group also does not have that tacit knowledge that individuals with more domain expertise in that particular area. For example, suppose someone encounters an unknown powder. An instinct and recommendation may be to identify it using the five senses. Still, those with more domain expertise may assume it is dangerous and contact the authority instead. To mitigate potential disparities, we encourage constructing focus groups for a variety of backgrounds to review new safety-related tasks and metrics. This would minimize incorrect assumptions and maximize coverage of the different types of covertly unsafe physical harm. 

\noindent \textbf{Crowd Sourcing.} 
Crowd workers are likely involved in many stages of the pipeline, from helping to write context to unsafe scenarios to human evaluation of the safety of generated texts. Like task creation, crowd workers may have unique perceptions of safety influenced by their backgrounds and beliefs \cite{sap2021annotators}. As a result, it is ideal to go beyond a simple convenience sample and acquire crowd workers with diverse perspectives to help mitigate biases that may span from perceptions of safety. For future research, this can be expanded to explore different definitions of safety.

\subsection{Bridging Gaps with Social Workers}
Social workers can bridge the gap between impacted communities, computer scientists, and policymakers. Since social workers are often immersed in marginalized communities \cite{mathiyazhagan2021social}, they can help computer scientists and policymakers understand different user groups and impacted communities, providing critical feedback on defining, measuring, and mitigating unsafe language from human-written or machine-generated text. Furthermore, social workers can help educate these communities to exercise caution when interacting with intelligent systems or machine learning models, as system outputs may not necessarily be truthful or safe. Social workers understand the cultural backgrounds of minority communities and can provide insight into misunderstandings or situations in which misinformation may be more likely to be accepted. A collaboration between domain experts and social workers can further benefit communities by advising on the risks of unsafe situations. 

\section{Conclusion}
In this paper, we address increasing concerns over text safety. We first establish degrees of separation with respect to physical harm as a methodology to label physically unsafe text as either overtly, covertly, or indirectly unsafe. We further dissect covertly unsafe text with the cause of either limited, incompatible, or incorrect information. Each type of covertly unsafe text has unique attributes requiring different strategies to resolve; we discuss these methods with respect to the ML pipeline to provide future researchers inspiration to tackle the issues of text safety. Finally, we discuss an interdisciplinary approach to mitigating covertly unsafe text. 

Covertly unsafe text is a challenging problem that spans a breadth of domains with no overtly unifying common thread. Since covertly unsafe text is subtle yet equally dangerous to overtly unsafe text, we argue that stakeholders and policymakers must acknowledge and proactively prioritize it to protect users' physical safety when interacting with intelligent systems.

\section*{Limitations}
While our research touches upon physical harm, our paper primarily discusses covertly unsafe text, limiting the discussion of other types of physically harmful text, including overtly unsafe and indirectly unsafe text. While the latter types of unsafe text are equally problematic in causing physical harm, our paper does not focus on either of these aspects due to the expansive coverage of previously existing research on these topics. 

In addition to limitations in the spectrum of physically harmful text, our work may be limited in categorizing covertly unsafe text. We provide subcategories of limited, incompatible, and incorrect information that causes text to be covertly unsafe, but these categories may not be comprehensive. 

This research aims to address the problem of covertly unsafe text and inspire future researchers to help improve intelligent systems by exploring ways to tackle this challenging problem. We encourage readers to consider the problem space of covertly unsafe text, whether there may be additional categorizations of these texts, and even propose new mitigation strategies. 

\section*{Ethical Considerations}\label{sec:ethics}
We acknowledge that our research touches upon sensitive topics of harm that affect individuals differently. Our work discusses commonsense and categorizations of harm with a singular definition of safety in an attempt to improve text safety universally, yet we note that personal backgrounds influence and shape people's views and values non-uniformly, which can affect people's perceptions of harm and safety differently. As a result, bias may propagate through efforts to improve text safety, which can impact protected groups disproportionately. We encourage researchers in this area to be aware of these potential factors and proactively attempt to mitigate bias against protected groups by applying a conscious human-centered strategy.

\section*{Acknowledgements}
We thank our reviewers for their helpful feedback. We also thank Rukmini Bapat for her early contributions in the initial literature search. We would also like to thank Amazon AWS Machine Learning Research Award and Amazon Alexa Knowledge for their generous support. This material is based upon work supported in part by the National Science Foundation under Grant \#2048122. The authors are solely responsible for the contents of the paper, and the opinions expressed in this publication do not reflect the official policy or position of the funding agencies. We also thank the Robert N. Noyce Trust for their generous gift to the University of California via the Noyce Initiative.


\bibliography{custom}
\bibliographystyle{acl_natbib}

\end{document}